\documentclass[runningheads]{llncs}
\usepackage{graphicx}
\usepackage{amsmath,amssymb} 
\usepackage{ruler}
\usepackage{color}
\usepackage{enumitem}

\usepackage{microtype}
\usepackage{graphicx}
\usepackage{subfigure}
\usepackage{booktabs} 
\usepackage{mathtools}
\usepackage{multirow}
\usepackage{enumitem}
\usepackage[detect-none]{siunitx}
\makeatletter
\@namedef{ver@everyshi.sty}{}
\makeatother
\usepackage{tikz}

\usetikzlibrary{fadings}
\pgfdeclareverticalshading{up}{100bp}
{color(0bp)=(pgftransparent!0); color(50bp)=(pgftransparent!0);
 color(75bp)=(pgftransparent!25); color(100bp)=(pgftransparent!25)}
\pgfdeclarefading{up}{\pgfuseshading{up}}    
\pgfdeclareradialshading{thick ring}{\pgfpointorigin}{
  color(0pt)=(pgftransparent!100); color(15pt)=(pgftransparent!100); 
  color(20)=(pgftransparent!0);
  color(25bp)=(pgftransparent!100); color(50bp)=(pgftransparent!100)}
\pgfdeclarefading{thick ring}{\pgfuseshading{thick ring}}    
\begin{tikzfadingfrompicture}[name=ball]
\shade [shading=ball, ball color=black] circle [radius=1];
\end{tikzfadingfrompicture} 
\begin{tikzfadingfrompicture}[name=ring]
\fill [white, path fading=thick ring] circle [radius=1];
\fill [black, path fading=up, fading angle=215] circle [radius=1];
\end{tikzfadingfrompicture}    

\tikzset{shiny ball/.style={
  fill=none, draw=none, shading=ball, shading angle=-15,
  postaction={fill=white, path fading=ball, opacity=0.75, fading angle=45},
  postaction={fill=white, path fading=ring}
}}

\begin{document}
\pagestyle{headings}
\mainmatter

\def\ACCV22SubNumber{***}  

\title{OOD Augmentation May Be at Odds with \\ Open-Set Recognition} 



\titlerunning{OOD Augmentation May Be at Odds with \\ Open-Set Recognition}
\authorrunning{M. Azizmalayeri, M. H. Rohban}

\author{Mohammad Azizmalayeri, Mohammad Hossein Rohban}
\institute{
m.azizmalayeri@sharif.edu \quad rohban@sharif.edu \\
Department of Computer Engineering, Sharif University of Technology, Tehran, Iran}

\maketitle

\begin{abstract}
Despite advances in image classification methods, detecting the samples not belonging to the training classes is still a challenging problem. There has been a burst of interest in this subject recently, which is called Open-Set Recognition (OSR). In OSR, the goal is to achieve both the classification and detecting out-of-distribution (OOD) samples. Several ideas have been proposed to push the empirical result further through complicated techniques. We believe that such complication is indeed not necessary. To this end, we have shown that Maximum Softmax Probability (MSP), as the simplest baseline for OSR, applied on Vision Transformers (ViTs) as the base classifier that is trained with non-OOD augmentations can surprisingly outperform many recent methods. Non-OOD augmentations are the ones that do not alter the data distribution by much. Our results outperform state-of-the-art in CIFAR-10 datasets, and is also better than most of the current methods in SVHN and MNIST. We show that training augmentation has a significant effect on the performance of ViTs in the OSR tasks, and while they should produce significant diversity in the augmented samples, the generated sample OOD-ness must remain limited.
\end{abstract}


\section{Introduction}
\label{Introduction}

\begin{figure}[ht]
\begin{center}
\centerline{\includegraphics[scale=0.46]{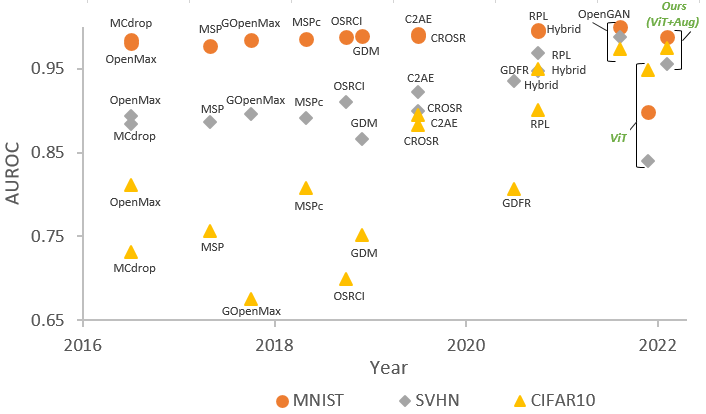}}
\caption{AUROC of recent state-of-the-art methods on MNIST, SVHN, CIFAR-10 sorted by year. Our method improves the MSP \cite{Hendrycks2017} via training vision transformers with augmentations that make fewer changes to the original data distribution to improve open-set recognition while retaining the augmentation diversity to reach high accuracy.}
\label{fig_comparison}
\end{center}
\vskip -0.2in
\end{figure}

Machine learning models have perfect generalization power even for some of the data they have not seen during training. Although this can improve model performance and increase test accuracy, it is hazardous in some cases, such as autonomous driving that needs to detect anomalous data and respond quickly. Therefore, it is necessary to provide methods that detect inputs that deviate significantly from the normal domain, also called ``out-of-distribution (OOD) inputs", without reducing the test accuracy.
This track of research has been explored extensively in various categories such as Novelty Detection (\emph{ND}) and One-Class Classification (\emph{OCC}). Specifically, in this work, we focus on Open-Set Recognition (OSR) that assumes a subset of classes as the known (closed) classes and the others as unknown (open) data. The model trained on known classes should also be able to discriminate between the known and unknown classes of data. The difficulty of OSR compared to the other categories is the closeness of known and unknown classes distribution in OSR as they belong to the same dataset. 

Previous methods on OSR can be divided into two main categories. The first group uses the likelihood extracted from the model itself. In this regard, the maximum of softmax probabilities (MSP) \cite{Hendrycks2017} and calibrated MSP (MSPc) \cite{Liang2018} are criteria that were proposed as baselines for OSR. OpenMax \cite{Bendale2016} uses the logits instead of softmax outputs to measure the distance of each sample to the average of each training class. The other track of works uses GANs. G-OpenMax \cite{Ge2017} is an extension of OpenMax that uses GANs to generate fake unknown images for training the classifier. OSRCI \cite{Neal2018} improves G-OpenMax by generating counter-factual images that are samples near the classifier decision boundaries. OpenGAN \cite{Kong2021} uses GANs with a different approach. It generates fake images (features) to train a discriminative model adversarially to distinguish the known classes from the fake ones. In addition to these two categories, there are also other methods such as C2AE \cite{Oza2019} and CROSR \cite{Yoshihashi2019} that measure an autoencoder reconstruction error for each sample, which could be used to discriminate open and closed classes. Currently, OpenGAN outperforms all the prior methods.

Clearly, research has progressed to more sophisticated methods to enhance the results. In this work, we show that the MSP method introduced as a baseline can perform better than all of them using Vision Transformer (\emph{ViT}) \cite{Dosovitskiy2021} as the classifier, but needs replacement of the standard training augmentations of ViTs with augmentations such as AutoAugment \cite{Cubuk2019} and RandAugment \cite{Cubuk2020}. More importantly, inspired from \cite{darmour20}, we believe that there are under-specifications in various ViT models for the OSR task. Specifically, ViT models with similar high classification accuracy could exhibit a huge variance in their OSR performance (see Fig. \ref{fig_scl}). To alleviate the problem, we suggest picking the training augmentations carefully, and also avoid supervised contrastive learning. Our results are compared with prior methods in Fig. \ref{fig_comparison}. Our method AUROC on CIFAR-10 datasets is higher than OpenGAN (state-of-the-art), and is also better than most of the current methods on SVHN and MNIST. These results are appealing since the method does not have any computational overhead for the model and does not face the problems of previous methods, such as unstable training of GANs. On the other hand, there are no limitations on combining it with previous methods.

\begin{figure*}[t]
\begin{center}
\centerline{\includegraphics[scale=0.43]{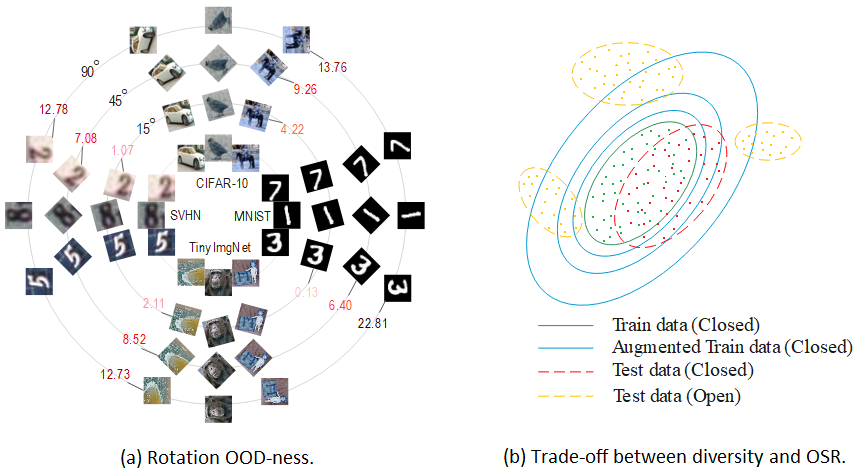}}
\caption{Effect of augmentation on open-set recognition. a) Rotation is a standard augmentation in vision tasks. We measured the difference between the distribution of augmented images and the main images based on our OOD-ness criterion for three different levels of rotations ($15^{\circ}$, $45^{\circ}$, and $90^{\circ}$) and displayed it for four different datasets. This difference increases with the rotation size, and augmented images will have a highly different distribution. b) Augmentation is used to generate more diverse images to avoid overfitting in training and reach a better test accuracy, but this can also lead to learning the open-set classes due to the changes in the original training distribution that prevent the model from detecting open-set samples. This problem is exacerbated by increasing the augmentation level, which is shown schematically.}
\label{fig_oodness}
\end{center}
\vskip -0.2in
\end{figure*}

ViTs are suitable for the OSR in that they have solved the problem of achieving satisfactory accuracy on the known data \cite{vaze2022openset}. This lets us to focus more on the problem of recognizing open classes. In this regard, we demonstrate that standard augmentations of training ViTs such as Rotation,  Flip, and MixUp \cite{Zhang2018} generate samples with high OOD-ness as displayed schematically in Fig. \ref{fig_oodness}. Therefore, we replace these augmentations with AutoAugment and RandAugment, which we demonstrate that suffer less from this problem. We also point out the importance of pre-training ViTs in OSR by comparing pre-trained models with FractalDB \cite{Nakashima2021} and ImageNet-1k \cite{Deng2009} data.

The significance of considering open classes in training ViTs is not limited to augmentation. Other training parts and algorithms also should be revisited. In this regard, Supervised Contrastive Learning (\emph{SCL}) \cite{Khosla2020} is a common learning algorithm that previous works have shown that performs better than standard training in tasks such as novelty or out-of-distribution detection \cite{Tack2020,Winkens2020}. We show that although the SCL contrasts the known classes from each other, it leads to the closeness of an open class distribution to one of the known classes distribution in the model feature space. This phenomenon can cause the model to assign open classes to the known ones. Therefore, we would not be able to discriminate open classes, which shows itself off well in ViTs with high classification capability.

In summary, our key contributions are the following:
\begin{itemize} [label={$\bullet$}, topsep=3pt, itemsep=3pt]
    \item Demonstrating the significance of considering the open classes in training on closed classes.
    
    \item Defining measures for OOD-ness and providing a framework to select better augmentations in training due to the trade-off between generalization and OSR.
    
    \item Considering ViTs for OSR and studying the effect of pre-training and model size on its performance.
    
    \item Outperforming all current methods via revisiting the training augmentations of ViTs, without any extra computational overhead.
    
    \item Demonstrating that contrary to previous works claims, the SCL can be harmful to OSR.
\end{itemize}

\section{Vision Transformer for Open-Set Recognition}
\label{ViT}

The success of transformers in the processing language models has led the vision community to move toward using them in image recognition. The superiority of transformers is mainly due to the attention to all parts of the input simultaneously \cite{Vaswani2017}. Following a similar idea in image recognition, ViTs were proposed \cite{Dosovitskiy2021}. The base transformer in ViT is similar to the language models but the 2D input image should be converted to 1D sequences to be compatible with transformers. Therefore, the 2D image $x \in \mathbb{R}^{C\times H\times W}$ is converted to 2D patches $x_p \in \mathbb{R}^{C\times P\times P}$ where $C$ is the channel numbers and ($H, W$) is the image size, and ($P, P$) is the patch size. Next, each patch is flattened and mapped to a $D$-dimentional embedding by a linear projection. These embeddings are called patch embeddings, which are padded with positional embedding to retain positional information and used as the transformer inputs.

In addition to the high capability of ViTs in image classification mentioned in the original paper, this model has also been compared to CNNs on some other problems and outperformed them. The study on the robustness of ViTs against adversarial perturbations indicated that they are more robust than CNNs and generalize well to unseen transformations since they learn more high-level features \cite{Shao2021,Naseer2021}. Another study introduced TransUNet that uses transformers for medical image segmentation and achieves superior performances than various competing methods due to the strong innate self-attention mechanism in transformers \cite{Chen2021}. Learning high-level features and attention mechanism can also be helpful in OSR due to the closeness of open and closed sets.

The advantages of ViTs over CNNs lead to their success in OSR, even with a simple method like MSP \cite{vaze2022openset}. In the following, we conduct experiments to indicate the superiority of ViTs in OSR and the effect of model size on its performance. We also point out the significance of the pre-training method. To this end, the settings of the experiment are stated first. After that, the results are analyzed. Note that due to the importance of training augmentation, its impact is investigated separately in the Section \ref{Augmentation}.

\subsection{Setup}
\label{Setup}

\quad\  \textbf{Models:} We consider the $3$ variants of the original ViT \cite{Dosovitskiy2021} including ViT-\{Tiny, Small, Base\}. This lets us compare the standard ViT model with previous methods and make a comparison on the impact of model size. Furthermore,  Data-efficient image Transformer with a Tiny size (\emph{DeiT-Ti}) \cite{Touvron2021} is also considered with the aim of comparing the impact of pre-training methods. DeiT alleviates the problem of requiring large-scale datasets for pre-training ViTs. 

\textbf{Pre-training:} The data used to pre-train ViT is an effective parameter on OSR. The similarity of pre-training data and the target dataset can let the model learn better features on closed classes, improving OSR. It can also let the model preserve the learned features on open class, which can harm OSR. Thus, the effect of pre-training data should be sifted through. To this end, we compare DeiT-Ti models pre-trained on well-known ImageNet-1k \cite{Deng2009}, FractalDB-1k, and FractalDB-10k \cite{Nakashima2021}. FractalDB pre-training attempts to pre-train models without natural images. Instead, it uses data generated by fractals, which is a formula-driven image dataset \cite{Kataoka2020}. Therefore, pre-training and target datasets share minimal information.

\textbf{Datasets:} To make the experiments more comprehensive, we consider three different datasets \{MNIST \cite{MNIST}, SVHN \cite{netzer2011reading}, and CIFAR-10 \cite{krizhevsky2009learning}\}, which are widely used in prior works \cite{Zhang2020,Perera2020}. These datasets contain $10$ classes. We randomly select $6$ classes as known (closed) classes and others as open ones in these datasets. Note that all experiments on each dataset are conducted over $5$ runs with different closed/open classes to reduce the effect of randomly selecting closed classes on the results.

\textbf{Training settings:} Following prior works, we fine-tune all models for $100$ epochs with the batch size $=128$, weight decay $=1e-4$, optimizer = SGD + momentum, learning rate $=0.01\times batch\_size/512$, and cosine learning rate schedule. Our only difference is that we use AutoAugment \cite{Cubuk2019} instead of standard ViT augmentation such as Rotation, Flip, and MixUp \cite{Zhang2018}. The reason for this change and its significance in results is discussed accurately in Section \ref{Augmentation}.

\textbf{OSR method:} A simple but effective method for OSR is Maximum Softmax Probability (\emph{MSP}) \cite{Hendrycks2017}. This method uses $max_{c\in\{1, 2, \ldots, k\}} f_c(x)$ as the score function that classifier $f$ returns for input $x$. This method has no computational overhead during the train/test, and can be used easily.

\textbf{Evaluation metric:} The Area Under the Receiver Operating Characteristic curve (\emph{AUROC}) is a well-known classification criterion that combines the True Positive Rate (\emph{TPR}) and False Positive Rate (\emph{FPR}) in a single metric. We use this criterion to determine the performance of each model in the binary classification of open vs. closed classes. The value of $0.5$ for AUROC means that the classifier assigns labels randomly, and the closer it is to $1$, the better the classifier performance.

\subsection{Experiments}
\label{Experiments}

\quad\ \textbf{Comparison with prior works:} We use ViT-B with the defined settings to compare our methods with the previous methods described in Section \ref{Introduction}. We also compare with the standard ViT-B trained with typical augmentation sets mentioned earlier as a baseline. The results are displayed in Fig. \ref{fig_comparison}. Based on these results, our method outperforms the standard ViT by a significant margin on all $3$ datasets. We investigate the reason of this improvement in detail in Section \ref{Augmentation}. Also, our method is better than all other methods on CIFAR-10 and outperforms OpenGAN \cite{Kong2021} as previous state-of-the-art (\emph{SOTA}). Note that OpenGAN overcomes the unstable training of the discriminator using a validation set containing the same open classes as the test dataset, which makes an explicit bias in their results. Moreover, while OpenGAN uses generative models to generate new OOD samples and discriminates them from the original ones with a GAN-discriminator, we only use MSP to discriminate open and closed samples without any extra computational overhead. On MNIST, almost all methods have satisfactory performance, and basically, the problem is solved. Our method is also satisfactory on SVHN, and  is among the $3$ best methods. Thus, the results are remarkably well on average, and the method is applicable to all the datasets.

\textbf{Impact of pre-training:} 
We believe that the remarkable superiority on CIFAR-10 is due to pre-training the classifier on a similar domain (ImageNet). To assess this hypothesis, and the effect of pre-training, we conducted experiments with the explained setting for pre-training in Section \ref{Setup}. The results are presented in Table \ref{tab_pretraining}. The pre-trained model on ImageNet-1k obviously outperforms the pre-trained model on FractalDB, while on MNIST and SVHN, they are almost the same. On the other hand, CIFAR-10 has much in common with ImageNet-1k, but MNIST and SVHN bear almost no resemblance to any of the data used for pre-training. This confirms our hypothesis.

\begin{table}[t]
\setlength{\tabcolsep}{3.5pt}
\caption{OSR AUROC of various pre-training methods of DeiT-Ti as the classifier. Pre-training on a similar domain to the target dataset has improved the AUROC for CIFAR-10. 
}
\label{tab_pretraining}
\begin{center}
\begin{small}
\begin{sc}
\begin{tabular}{lccc}
\toprule
\multirow{2.5}{*}{Dataset} &    \multicolumn{ 3}{c}{Pre-training} \\

\cmidrule{2-4}
\multicolumn{ 1}{c}{} &     Fractal-$1$k & Fractal-$10$k & ImgNet-$1$k \\
\midrule
MNIST      & $0.983$      & $0.980$        & $0.981$   \\
SVHN       & $0.937$      & $0.944$       & $0.940$    \\
CIFAR-10    & $0.818$      & $0.820$        & $0.900$  \\
\bottomrule
\end{tabular}
\end{sc}
\end{small}
\end{center}
\vskip -0.2in
\end{table}

\begin{figure}[t]
\begin{center}
\centerline{\includegraphics[scale=0.55]{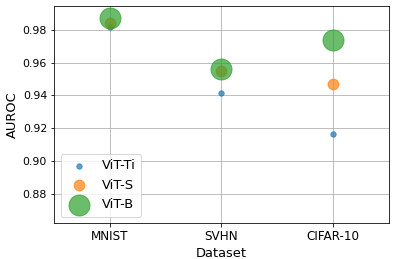}}
\caption{Effect of model size on OSR. ViT-\{Ti, S, B\} are used as the main model and their performance on OSR is evaluated on $4$ datasets using AUROC.}
\label{fig_size}
\end{center}
\vskip -0.2in
\end{figure}

\textbf{Impact of model size:} To investigate the effect of model size on the OSR, we use the $3$ variants of the original ViT. ViT-Tiny is used as the smallest model with only $5.7$ million parameters and ViT-B as the largest model with $86$ million. These model are compared in Fig. \ref{fig_size}. The results show that the larger the model size, the higher the AUROC value, and the better the model can discriminate between open and closed classes. This is due to the fact that larger ViT can learn more robust features that let the model detect OOD samples \cite{Shao2021,Mahmood2021}. Of course, in MNIST and SVHN, the difference between ViT-B and ViT-S is insignificant. This may be due to the simplicity of the data that even small models can learn robust features on it \cite{Madry2018}.

\begin{figure}[t]
\begin{center}
\centerline{\includegraphics[scale=0.35]{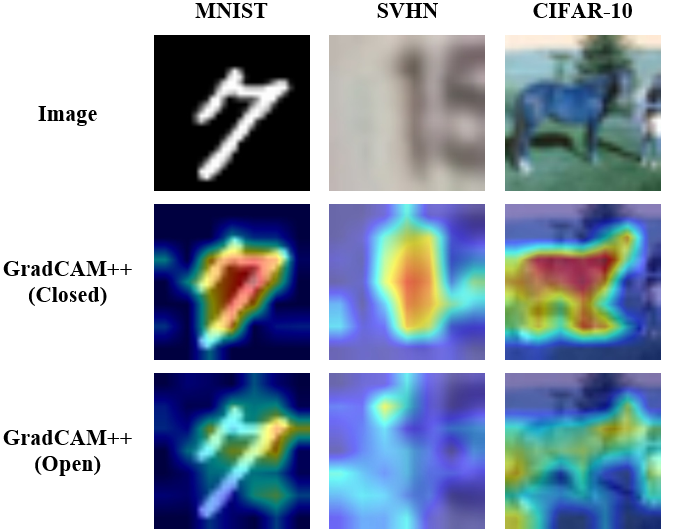}}
\caption{Visualization of the attention map on images from various datasets in cases considered among the open or closed classes in training using GradCAM++. More or less attention is represented with red and blue colors, respectively.}
\label{fig_gradcam}
\end{center}
\vskip -0.2in
\end{figure}

\textbf{Attention analysis:} Attention is a key element for achieving high robustness in image classification. The attention map on an image indicates the main parts of the image involved in assigning the label to the image. ViTs use attention mechanism that enables the model to extract the local and global features in an image. As a result, they can pay proper attention to the main parts of the images. In this regard, a key factor that can help the model discriminate the OOD samples is the attention map. It is expected that the model would not be able to have proper attention on open classes as much as closed ones. To investigate this, we use Grad-CAM++ \cite{Chattopadhyay2018} to visualize the attention maps on some images in both cases that are considered as open or closed classes. GradCAM++ is the improved version of GradCAM \cite{Selvaraju2017} that uses the gradients flowing into the final layers of the model to highlight important regions in the image. The results are displayed in Fig. \ref{fig_gradcam}. In some images, the model has utterly wrong attention on the open classes, which is reasonable since they are not seen in training. Of course, in some open images, the model has the proper attention (not as good as the closed case) due to the similar domain of open and closed classes that let the model extract features even on open classes.

\section{Augmentation OOD-ness}
\label{Augmentation}

Data augmentation is mainly used to prevent machine learning models from learning irrelevant patterns by representing an extended version of the training data. Augmentation helps the model to generalize well on validation and test sets \cite{Perez2017}. Selecting the type of augmentations are also an important issue, e.g., better augmentations in adversarial training (\emph{AT}) led to huge improvement in robust model than complicated modifications of AT \cite{rebuffi2021}. Thus, the ability of augmentation to prevent overfitting has led to their widespread use in training, but we should revisit the effect of augmentation on OSR.

Despite the benefits of augmentations, they can shift the distribution of data. Changes in the training data distribution can allow the model to learn features related to open classes \cite{He2019}. This concern is displayed in Fig. \ref{fig_oodness}. Since we can not eliminate augmentation due to overfitting, we must move forward to use augmentations whose distribution of augmented data is not much different from the distribution of the original data. We call this difference the ``Augmentation OOD-ness," since its large value means that augmented data are like the OOD data compared to the original data distribution.

In the following, we first introduce some criteria to measure the diversity and OOD-ness of augmentations. Next, we use them to compare various augmentations' effects on OSR. Finally, we empirically validate our claims by measuring the effect of augmentations on OSR results.

\subsection{Comparison criteria}\label{Comparison_criteria}
To compare the augmentations, we introduce Diversity as a measure that considers overfitting, and we define two measures to consider the OOD-ness.

\textbf{Diversity:} The impact of an augmentation on the generalization on the test set depends on the variety that it produces in the training data. This variety can be measured with the Diversity criteria \cite{Gontijo2021}. Diversity 
is defined as follows:
\begin{equation}\label{diversity}
    \mathcal{D}[\mathcal{A}, m, D_{train}] = \mathbb{E}_{\mathcal{A}(D_{train})} [L_{m}] / \mathbb{E}_{D_{train}} [L_m],
\end{equation}
where $L_m$
is the loss of the model $m$
that is trained 
with the augmentation $\mathcal{A}$.
A higher value of $\mathcal{D}$ indicates a more difficult model fit to the augmented data, which in turns shows higher diversity resulting from the augmentation. We use the original paper results in our comparisons.

\textbf{OOD-ness:} Following the OOD-ness definition in \cite{Tack2020} and the Diversity above, we define two new OOD-ness measures specifically designed to find augmentations that generate OOD outputs. To define the first criteria, we assume that, unlike the definition of Diversity, the model $m$ is trained {\it without} any augmentations. This way, we can measure whether the model can work with a similar performance on an unknown augmented data or not. With this assumption, we use the same equation as Eq. \ref{diversity} as the first criterion for the OOD-ness. Here, the difference between the distribution of augmented and original data is measured, while in Diversity the ability of the model to fit to the augmented data is considered.

As the second criterion, we first assume the original data as the in-distribution and the augmented data as the OOD. Next, we use MSP as score function to calculate the AUROC under this assumption. Suppose the value of AUROC is close to $1$. In that case, the assumption is correct, and the augmented and original data are separated, but the closer it is to $0.5$, the assumption is more inaccurate, and both come from the same distribution. Thus, the AUROC value can be considered as a measure for OOD-ness.

\subsection{Experiments}
For the experiments in this section, $11$ different augmentations are compared in terms of Diversity and OOD-ness using DeiT-Ti pre-trained with FractalDB-10k to eliminate the effect of pre-training augmentations on the results as much as possible. The details of the augmentations are available in Appendix \ref{augmntation_details}. Moreover, the experiments are conducted on MNIST, SVHN, and CIFAR-10.

\begin{table}[t]
\setlength{\tabcolsep}{3.5pt}
\renewcommand{\arraystretch}{1.15}
\caption{OOD-ness for different augmentations measured with our defined criteria.}
\label{tab_oodness}
\begin{center}
\begin{small}
\begin{sc}
\begin{tabular}{l|cc|cc|cc}
    \toprule
    Dataset  & \multicolumn{2}{c|}{MNIST} & \multicolumn{2}{c|}{SVHN} & \multicolumn{2}{c}{CIFAR-10} \\
    \midrule
    OOD-ness & (A)   & (B)   & (A)   & (B)   & (A)   & (B) \\
    \midrule
    Cutout  & $33$    & $0.55$  & $104$   & $0.52$  & $382$   & $0.60$ \\
    Colorjit  & $9$     & $0.59$  & $209$   & $0.52$  & $2186$  & $0.67$ \\
    Noise & $298$   & $0.75$  & $1490$  & $0.66$  & $3468$  & $0.75$ \\
    AutoAug & $2864$  & $0.79$  & $1318$  & $0.62$  & $5030$  & $0.73$ \\
    RandAug  & $2760$  & $0.80$   & $2677$  & $0.69$  & $6000$  & $0.78$ \\
    Flip  & $10669$ & $0.78$  & $7072$  & $0.76$  & $3318$  & $0.69$ \\
    Rotate  & $9973$  & $0.83$  & $7213$  & $0.89$  & $8371$  & $0.90$ \\
    Permute  & $13139$ & $0.87$  & $12195$ & $0.87$  & $7987$  & $0.82$ \\
    MixUp  & $25269$ & $0.80$   & $14316$ & $0.73$  & $21286$ & $0.76$ \\
    FGSM  & $2350$  & $0.91$  & $12024$ & $0.97$  & $36977$ & $0.97$ \\
    PGD   & $9432$  & $0.92$  & $44837$ & $0.98$  & $76571$ & $0.97$ \\
    \bottomrule
\end{tabular}%
\end{sc}
\end{small}
\end{center}
\vskip -0.2in
\end{table}

\textbf{Measuring OOD-ness:} As mentioned above, diversity results are analyzed in the original paper. Therefore, we only measure the OOD-ness results using two criteria defined in Section \ref{Comparison_criteria}. The results are available in Table \ref{tab_oodness}. Both criteria are almost in agreement, and if an augmentation has a relatively high OOD-ness according to one, the other criterion can also show this. Based on these results, we categorize the augmentations into $3$ groups based on their OOD-ness: low-OOD (Colorjitter and Cutout), moderate-OOD (Noise, AutoAugment, and RandAugment), and high-OOD (Flip, Rotate, Permute, MixUp, FGSM, PGD). 

\textbf{OSR and generalization trade-off:} In normal training, one may utilize various augmentations such as rotation and flip with the aim of diversity and preventing overfitting, or transforms such as FGSM or PGD to reach a robust model \cite{Madry2018} without any other concerns. However, as indicated, some of these transforms suffer from OOD-ness and have the risk of generating OOD samples, which should be avoided for OSR. Thus, we can say that there is a trade-off between augmentations that are employed for generalization and OSR. Since both the generalization to test set and OSR are vital for us, we use Diversity and OOD-ness to compare all the augmentations in order to find ones that fit both. The results are presented in Table \ref{tab_tradeoff}. The mentioned compromise can also be deduced from this table because a high risky value for Diversity/OOD-ness is associated with a low or moderate risky value for OOD-ness/Diversity. Still, some of them are better than the others. If OOD-ness is prioritized, Cutout is better than the others since it has a moderate diversity. In case Diversity is prioritized, AutoAugment and RandAugment are better due to the moderate OOD-ness. Nevertheless, we still have to answer which of the mentioned cases is the best. In the following, we empirically investigate some of these augmentations directly in OSR to evaluate our claims and find the best possible choice.

\newcommand{\cirr}{\tikz\path [shiny ball, ball color=red] (0, 0) circle [radius=0.2];}
\newcommand{\cirg}{\tikz\path [shiny ball, ball color=green] (0, 0) circle [radius=0.2];}
\newcommand{\ciro}{\tikz\path [shiny ball, ball color=yellow] (0, 0) circle [radius=0.2];}
\begin{table}[t]
\setlength{\tabcolsep}{4.0pt}
\caption{Trade-off between OOD-ness (OOD) and Diversity (Div). The green, yellow, and red balls indicate the low, moderate, and high risks, respectively. High Risk for OOD-ness is equivalent to high values (distribution shift) and for Diversity is equivalent to low values (overfitting).}
\label{tab_tradeoff}
\begin{center}
\begin{small}
\begin{sc}
\begin{tabular}{lcc||lcc}
    \toprule
    Transform & OOD & Div & Transform & OOD & Div\\
    \midrule
    Cutout  & \cirg     & \ciro     & Flip  & \cirr     & \cirr \\
    Colorjit  & \cirg     & \cirr     & Rotate  & \cirr     & \cirr \\
    Noise & \ciro     & \ciro     & Permute   & \cirr     & \ciro \\
    AutoAug & \ciro     & \cirg     & MixUp & \cirr     & \cirg \\
    RandAug & \ciro     & \cirg     & FGSM, PGD & \cirr     & \cirg \\
    \bottomrule
\end{tabular}
\end{sc}
\end{small}
\end{center}
\vskip -0.2in
\end{table}

\textbf{Evaluating augmentations:} To evaluate the effect of augmentations directly on OSR, we select $5$ different augmentations to cover various aspects. Cutout is selected to prioritize OOD-ness while considering Diversity. RandAugment and AutoAugment are selected separately to prioritize Diversity while considering OOD-ness. The combination of Flip \& Rotation is selected to consider a set of standard augmentation with a bad Diversity and OOD-ness. Furthermore, to analyze the impact of high OOD-ness without any concern about Diversity, the combination of Flip \& Rotation \& AutoAugment is also considered as a set of augmentations. All of these augmentations are separately utilized for training the models on $3$ different datasets. The OSR performance for each model is reported in Table \ref{tab_aug_osr}. In the following, we investigate the results of each augmentation.

\begin{table*}[t]
\setlength{\tabcolsep}{3.0pt}
\caption{OSR AUROC performance of models trained with different augmentation sets using MSP on $3$ different datasets. The best and second-best results are distinguished with bold and underlined text for each dataset, respectively.}
\label{tab_aug_osr}
\begin{center}
\begin{small}
\begin{sc}
    \begin{tabular}{lccccc}
    \toprule
    \multirow{3}{*}{Dataset} &    \multicolumn{ 5}{c}{Augmentation} \\

\cmidrule{2-6}
\multicolumn{ 1}{c}{} & \multirow{2}{*}{Cutout} & \multirow{2}{*}{RandAug} & \multirow{2}{*}{AutoAug} & Flip \&  & Flip \& Rotate \& \\
    \multicolumn{1}{c}{}   &    &     &    & Rotate & AutoAug  \\
    \midrule
    MNIST & $\mathbf{0.985}$ & $0.974$ & \underline{$0.980$}  & $0.957$ & $0.946$ \\
    SVHN & $0.942$ & $\mathbf{0.947}$ & \underline{$0.944$} & $0.899$ & $0.866$ \\
    CIFAR-10 & $0.768$ & \underline{$0.805$} & $\mathbf{0.820}$ & $0.746$ & $0.795$ \\
    \bottomrule
    \end{tabular}%
\end{sc}
\end{small}
\end{center}
\vskip -0.2in
\end{table*}

\begin{itemize}[label={$\bullet$}, topsep=0pt, itemsep=2pt, leftmargin=10pt, parsep=0pt]
    \item Cutout: This augmentation reaches the best AUROC on MNIST and is not much different from the best result on SVHN but does not work well on CIFAR-10. These results are entirely consistent with our claims because in the first two datasets, especially MNIST, which do not suffer much from overfitting and the accuracy is almost the same for all augmentations, the best results are achieved with Cutout as the augmentation with the lowest OOD-ness. On the other hand, in CIFAR-10 the effect of overfitting is  significant, and the relatively low diversity of Cutout manifests itself that prevents the model from achieving good results.
    \item RandAugment/AutoAugment: Both augmentations reach competitive AUROCs on all datasets. This demonstrates that considering OOD-ness along with Diversity is essential for OSR. This is the reason that we used AutoAugment for the experiments in Section \ref{ViT}.
    \item Flip \& Rotation: The OSR performance of this augmentation set is not as good as the previous ones. This is due to the high OOD-ness and low diversity of these augmentations that make them the worst choice for training the model with the aim of OSR.
    \item Flip \& Rotation \& AutoAugment: The difference between this augmentation set and the previous one is in AutoAugment. Thus, this set has a higher OOD-ness and Diversity than the previous set. Considering Table \ref{tab_aug_osr}, we see that these points can also be deduced from the results. On MNIST and SVHN with minor overfitting problems and almost the same accuracy on closed classes, OOD-ness prevails, which worsens the results. Furthermore, on CIFAR-10 with more overfitting concerns, Diversity manifests itself and leads to a better AUROC. These results also support our claims about the trade-off between OOD-ness and Diversity.
\end{itemize}
As a result of these experiments, we can conclude that in the general case, AutoAugment and RandAugment are the best choices for OSR. Of course, if one ensures that there is a negligible overfitting issue on a specific data, Cutout would be a better option.

\textbf{Cross-dataset detection:} To show that the results can also be used within a large dataset, we used a cross-dataset OSR scheme. In this experiment, ImageNet-1k is considered as closed data and \{Places365 \cite{zhou2017places}, LSUN \cite{yu2015lsun}, and iSUN\cite{xu2015turkergaze}\} as open datasets. The results are presented in the Table \ref{tab_cross_dataset} that demonstrate the effectiveness of our proposed method to select a proper augmentation in training in comparison with OpenGAN and pure ViT.

\begin{table*}[t]
\caption{AUROC for cross-dataset OSR scheme with ImageNet-1k as closed dataset.}
\label{tab_cross_dataset}
\begin{center}
\begin{small}
\begin{sc}
\begin{tabular}{cccc}
\toprule
\multirow{2.5}{*}{Method} &    \multicolumn{ 3}{c}{Out Datasets} \\

\cmidrule{2-4}
\multicolumn{ 1}{c}{} &     Places &       LSUN &       iSUN \\
\midrule
   OpenGAN-fea &      0.578 &      0.665 &      0.661 \\

       ViT-B &      0.672 &      0.848 &      0.836 \\

   ViT-B + Aug &      0.796 &       0.920 &      0.915 \\
\bottomrule
\end{tabular} 
\end{sc} 
\end{small}
\end{center}
\vskip -0.2in
\end{table*}

\section{Contrastive Learning May Be Harmful}
\label{Contrastive}

The power of ViTs in closed classes classification allowed us to concentrate more carefully on open classes with less concern about closed ones. Consequently, we demonstrated that training augmentation significantly affects OSR. In addition to augmentation, it is necessary to review other parts of the training to evaluate their influence on open test classes. We use contrastive learning as a common training method to show the importance of attention to this issue.

Contrastive learning (\emph{CL}) is widely used in OOD detection due to the capability of learning diverse representation, e.g., CSI \cite{Tack2020} trains a confidence calibrated classifier \cite{Lee2018b} by adapting it to the Supervised Contrastive Learning (\emph{SCL}) \cite{Khosla2020} for Novelty Detection. As another example, Masked CL \cite{Cho2021} proposes an extension of CL with class-conditional mask and stochastic positive attraction that can shape class-conditional clusters by inheriting advantages of CL for anomaly detection.

In the following, we first express how the contrastive loss function works. We then discuss that this loss function can reduce OSR performance. Finally, we support our claims by visualizing the feature space and measuring the OSR performance for models trained with SCL.

\subsection{Contrastive training}

\textbf{Loss function:} CL (e.g., SimCLR \cite{Chen2020b}) learns the embedding representation by maximizing the similarity of features learned for different views of a single image while repelling them from other images in the batch. To this end, it assumes different augmentations of an image $x$ as the positive set $x_+$ and the other images as the negative set $x_-$. The similarity of the images in feature space is measured with cosine similarity as $sim(z, z')=z^\top z'/(||z||\ ||z'||)$. Then, the contrastive loss that aims to maximize $sim(z, z_+)$ and minimize $sim(z, z_-)$ is defined as:
\begin{equation} \label{contr}
    \mathcal{L}(x) = \frac{-1}{|x_+|}\ \sum_{x' \in x_+} \log \ \frac{exp(sim(z, z')/\tau)}{\sum\limits_{x'' \in (x_+\cup x_-)} exp(sim(z, z'')/\tau)},
\end{equation}
where $\tau$ is a positive temperature parameter and \{$z, z', z''$\} are the latent vectors for \{$x, x', x''$\}, respectively. SCL is a variant of CL that was proposed to include the images with the same label as $x$ in the positive set instead of negatives. This causes the model to learn  similar representations for different views of samples of a class.

\textbf{Contrastive risk in OSR:} Despite the advantages of SCL in learning better representation for closed classes, it has some consequences on open classes that impairs the OSR performance. Suppose input $x$ belongs to an open class. Although $x$ does not belong to any closed classes, it is probably more similar to some training instances. Hence, it gets closer to that class in embeddings and gets far from the others since we have taught the model so in the objective function of SCL. 
This makes it hard for the model to discriminate the open classes.

\subsection{Experiments}

\begin{figure}[t]
\begin{center}
\centerline{\includegraphics[scale=0.4]{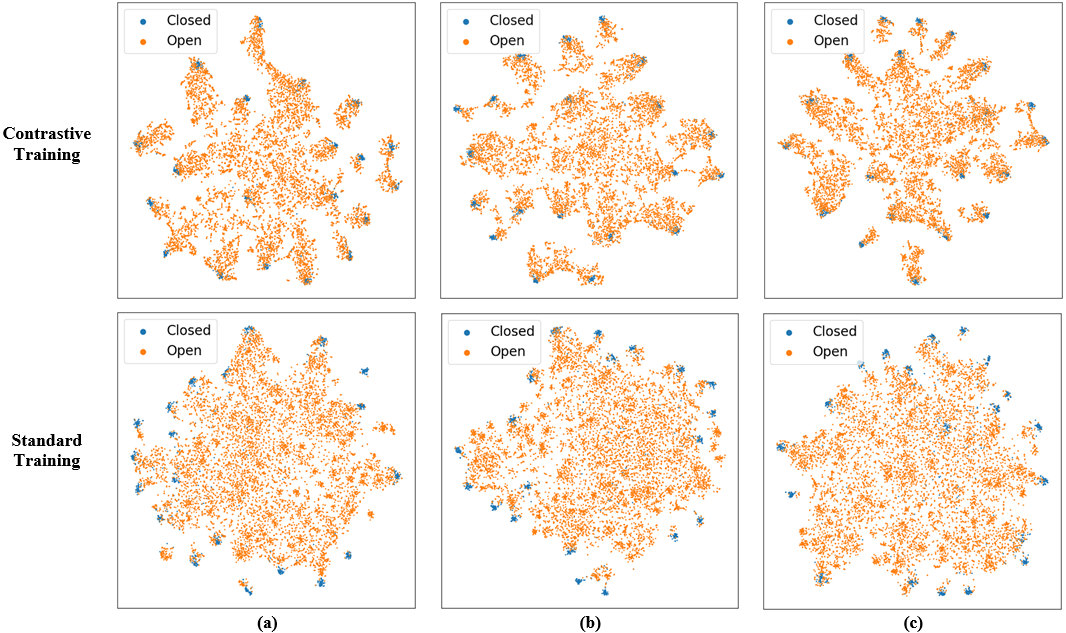}}
\caption{t-SNE visualization of features learned by ViT-B in standard and contrastive training for $3$ different closed and open classes on TinyImageNet. Contrastive training causes open classes to lean toward the closed classes.}
\label{fig_tsne}
\end{center}
\vskip -0.3in
\end{figure}

\textbf{Open-Set visualization:} To clarify the risk of SCL in OSR, we first train models with both SCL and standard methods on $180$ classes of TinyImageNet. This dataset has $200$ classes that makes it a good fit to investigate the impact of SCL on open classes. Next, t-SNE is used to visualize the latent space. The results are displayed in Fig. \ref{fig_tsne} for $3$ choices of closed classes. Based on these plots, the embedding of open classes lean toward the closed ones in SCL, unlike standard training. Obviously, the open classes in SCL are grouped next to the closed ones, while they are distributed uniformly in the standard training.

\begin{figure}[t]
\begin{center}
\centerline{\includegraphics[scale=0.55]{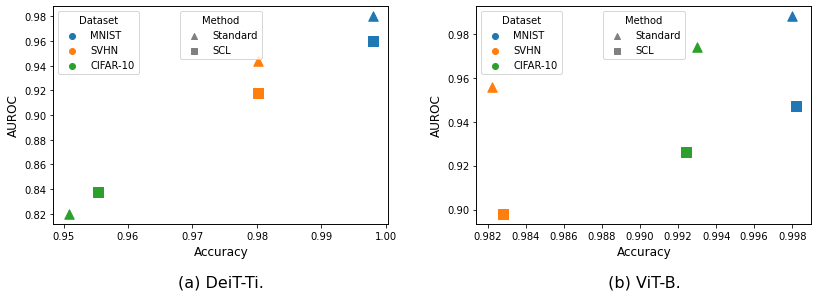}}
\caption{Comparison of standard and contrastive training in AUROC for OSR and accuracy on closed classes using DeiT-Ti pre-trained on FractalDB-10k and ViT-B pre-trained on ImageNet-1k in $3$ different datasets.}
\label{fig_scl}
\end{center}
\vskip -0.25in
\end{figure}

\textbf{Numerical evaluation:} To compare SCL and standard training more comprehensively, we train DeiT-Ti pre-trained with FractalDB-10k and ViT-B pre-trained with ImageNet-1k using both training methods on $3$ different datasets. This lets us consider the impact of model size and dataset in the comparisons.
The results are available in Fig. \ref{fig_scl}. The standard training outperforms SCL in all cases except in training DeiT-Ti on CIFAR-10. This indicates the harmful impact of SCL on OSR. SCL advantages only prevail when the model has a lower capability in classifying the closed classes (accuracy) that demonstrates the trade-off between the drawbacks and advantages of SCL.

The self-supervised methods that do not leverage negative samples (e.g., BYOL \cite{Bastien2020} and SimSiam \cite{Chen2021b}) may be possible solutions to this issue since an open sample would not be pushed toward a specific class by heterogeneous classes. However, these methods still can not compete with the supervised methods in OSR due to lower capability in classifying the closed classes.

\section{Conclusion}
\label{Conclusion}
ViTs have achieved excellent results in classifying closed classes, making them a good choice for OSR. To be competitive in OSR, two main requirements are essential. The first is an appropriate classifier for closed classes, which ViT does well. Second, the classifier should be able to distinguish the open and closed classes. Standard training of ViTs causes issues with the second requirement.  We noted that the training augmentation is playing a key role here. We analyzed the Diversity and OOD-ness of different augmentations in this work to find the best augmentation choices for OSR, which found to be AutoAugment. Using this finding, we could outperform the SOTA on CIFAR-10. We also demonstrated that pre-training and model size significantly affect the OSR performance. Finally, we used contrastive learning to demonstrate that other training techniques should also be revisited to consider their impacts on open classes independently of the closed ones. We hope that this work will be a beginning for future efforts to examine the training impacts on open classes.

\nocite{Gal2016}
\nocite{Lee2018}
\nocite{Chen2020}

\bibliographystyle{splncs}
\bibliography{main}


\appendix

\section{Augmentation details}
\label{augmntation_details}
In this section, we will express the details of augmentations used in section \ref{Augmentation} to measure the OOD-ness. Furthermore, we will also examine the impact of changing the augmentation level. To this end, $3$ different levels (Low, Moderate, and High) for each augmentation are specified to measure their OOD-ness. Also, we call the augmentations in Table \ref{tab_oodness}  as Mixture level since they are selected such that they include different levels.

Different augmentation levels are selected as follows:
\begin{itemize}[label={$\bullet$}, topsep=3pt, itemsep=3pt]
    \item Cutout: $N$ squares with size $S$ are cut out from the image with width $W$ and length $L$. For low, moderate, and high levels, $S=(W, L)/7$ and $N=1, 2, 4$ are used, respectively. For the mixture level, $N$ is selected randomly from the the range of $\numrange{1}{4}$ and $S=(W, L)/\alpha$ where $\alpha$ is selected in the range of $\numrange{7}{14}$.
    
    \item Colorjitter: Randomly jitters brightness, saturation, and contrast with factors in the range of $(1-\alpha)$ to $(1+\alpha)$, and hue with factor in the range of  $-\alpha$ to $\alpha$, where $\alpha$ can be set with different values. $\alpha=0.1, 0.3, 0.5$ is used for low, moderate, and high levels, respectively. For the mixture level, $\alpha$ for each parameter is selected randomly in the range of $\numrange{0.0}{0.5}$.
    
    \item Noise: Gaussian noise with standard deviation $\alpha$ is added to the image. $\alpha=0.1, 0.25, 0.5$ is used for low, moderate, and high levels, respectively. For the mixture level, $\alpha$ is selected randomly in the range of $\numrange{0.0}{0.5}$.
    
    \item AutoAugment: We use $3$ different policies learned on SVHN, CIFAR-10, and ImageNet datasets as low, moderate, and high levels. The policy is selected randomly from the mentioned policies for the mixture level.
    
    \item RandAugment: $N$ transformations are applied sequentially with magnitude $\alpha \in [0, 1]$. For low, moderate, and high levels, $(N,\alpha)$ = $(2,\ 0.125),$ $(4,\ 0.25),$ $(8,\ 0.5)$ are used, respectively. $(N,\alpha)$ for mixture level is selected randomly from the mentioned values.
    
    \item Flip: Horizontal and vertical flips and their combination are used as low, moderate, and high levels, respectively. For the mixture level, horizontal and vertical flips are applied with a probability of 0.5.
    
    \item Rotate: The images are rotated by $15^{\circ}$, $45^{\circ}$, and $90^{\circ}$ for low, moderate, and high levels, respectively. Amount of rotation for mixture level is selected randomly from the range of $-90^{\circ}$ to $90^{\circ}$.
    
    \item Permute: For the low level, the image is split from length into two parts and replaced with each other. Similarly, the image is split from width for the moderate level. Each of these transforms is applied with the probability of 0.5 for mixture level, and both are applied for high level.
    
    \item MixUp: Images in a batch are combined with weights sampled from $Beta(\alpha, \alpha)$ distribution. $\alpha=0.1, 0.5, 1.0$ is used for low, moderate, and high levels, respectively. For the mixture level, $\alpha$ is selected randomly in the range of $\numrange{0.0}{1.0}$.
    
    \item FGSM, PGD: These transforms are not commonly used as augmentation, but they are widely used to reach a robust model against adversarial perturbations. Therefore, we also considered them in our investigations. These transforms add a perturbation to the image with $\ell_\infty$ norm less than $ \epsilon$. We use $\epsilon = 2/255, 4/255, 8/255$ for low, moderate, and high levels, respectively. For the mixture level, $\epsilon$ is selected randomly in the range of $0.0$ to $8/255$.
\end{itemize}

As mentioned earlier, OOD-ness for mixture level is reported in Table \ref{tab_oodness}. For other levels, OOD-ness on MNIST, SVHN, and CIFAR-10 datasets are provided in Tables \ref{tab_oodness_mnist}, \ref{tab_oodness_svhn}, and \ref{tab_oodness_cifar10}, respectively. We hope that these details and results help to understand the augmentations OOD-ness better.

\begin{table*}[t]
\setlength{\tabcolsep}{4.0pt}
\renewcommand{\arraystretch}{1.2}
\caption{OOD-ness for different augmentations with $3$ different levels on MNIST.}
\label{tab_oodness_mnist}
\begin{center}
\begin{small}
\begin{sc}
\begin{tabular}{l|cc|cc|cc}
    \toprule
    Aug Level & \multicolumn{2}{c|}{Low} & \multicolumn{2}{c|}{Moderate} & \multicolumn{2}{c}{High} \\
    \midrule
    OOD-ness & (A) & (B) & (A) & (B) & (A) & (B) \\
    \midrule
    Cutout  & $25$ & $0.53$ & $59$ & $0.56$ & $177$ & $0.62$ \\
    Colorjit  & $1$ & $0.52$ & $11$ & $0.6$ & $38$ & $0.69$ \\
    Noise & $2$ & $0.55$ & $48$ & $0.72$ & $2014$ & $0.99$ \\
    AutoAug & $3832$ & $0.77$ & $2861$ & $0.73$ & $1838$ & $0.83$ \\
    RandAug & $509$ & $0.64$ & $1576$ & $0.8$ & $6172$ & $0.96$ \\
    Flip  & $10215$ & $0.86$ & $12761$ & $0.87$ & $19440$ & $0.89$ \\
    Rotate  & $138$ & $0.58$ & $6409$ & $0.90$ & $22814$ & $0.98$ \\
    Permute  & $24935$ & $1.00$ & $11513$ & $0.99$ & $16044$ & $1.00$ \\
    MixUp  & $28765$ & $0.74$ & $24737$ & $0.81$ & $22245$ & $0.85$ \\
    FGSM & $235$ & $0.89$ & $1756$ & $0.97$ & $6433$ & $0.99$ \\
    PGD  & $351$ & $0.91$ & $4576$ & $0.99$ & $35705$ & $1.00$ \\
    \bottomrule
    \end{tabular}%
\end{sc}
\end{small}
\end{center}
\vskip -0.2in
\end{table*}

\begin{table*}[t]
\setlength{\tabcolsep}{4.0pt}
\renewcommand{\arraystretch}{1.2}
\caption{OOD-ness for different augmentations with $3$ different levels on SVHN.}
\label{tab_oodness_svhn}
\begin{center}
\begin{small}
\begin{sc}
    \begin{tabular}{l|cc|cc|cc}
    \toprule
    Aug Level & \multicolumn{2}{c|}{Low} & \multicolumn{2}{c|}{Moderate} & \multicolumn{2}{c}{High} \\
    \midrule
    OOD-ness & (A) & (B) & (A) & (B) & (A) & (B) \\
    \midrule
    Cutout  & $54$ & $0.51$ & $133$ & $0.53$ & $397$ & $0.58$ \\
    Colorjit  & $36$ & $0.51$ & $277$ & $0.53$ & $478$ & $0.54$ \\
    Noise & $131$ & $0.51$ & $1162$ & $0.61$ & $4405$ & $0.96$ \\
    AutoAug & $1080$ & $0.61$ & $1587$ & $0.60$ & $1073$ & 0.63 \\
    RandAug & $248$ & $0.53$ & $1185$ & $0.65$ & $6879$ & $0.90$ \\
    Flip  & $7600$ & $0.84$ & $11030$ & $0.83$ & $9236$ & $0.86$ \\
    Rotate  & $1073$ & $0.68$ & $7083$ & $0.99$ & $12780$ & $1.00$ \\
    Permute  & $12757$ & $1.00$ & $19820$ & $0.98$ & $15810$ & $0.99$ \\
    MixUp  & $15048$ & $0.71$ & $14241$ & $0.74$ & $13651$ & $0.75$ \\
    FGSM & $10817$ & $0.99$ & $14344$ & $0.99$ & $15353$ & $0.99$ \\
    PGD  & $32705$ & $1.00$ & $54364$ & $1.00$ & $64164$ & $1.00$ \\
    \bottomrule
    \end{tabular}%
\end{sc}
\end{small}
\end{center}
\vskip -0.2in
\end{table*}

\begin{table*}[t]
\setlength{\tabcolsep}{4.0pt}
\renewcommand{\arraystretch}{1.2}
\caption{OOD-ness for different augmentations with $3$ different levels on CIFAR-10.}
\label{tab_oodness_cifar10}
\begin{center}
\begin{small}
\begin{sc}
    \begin{tabular}{l|cc|cc|cc}
    \toprule
    Aug Level & \multicolumn{2}{c|}{Low} & \multicolumn{2}{c|}{Moderate} & \multicolumn{2}{c}{High} \\
    \midrule
    OOD-ness & (A) & (B) & (A) & (B) & (A) & (B) \\
    \midrule
    Cutout  & $144$ & $0.57$ & $507$ & $0.63$ & $1661$ & $0.72$ \\
    Colorjit  & $533$ & $0.57$ & $2722$ & $0.70$ & $4484$ & $0.76$ \\
    Noise & $105$ & $0.57$ & $2907$ & $0.79$ & $9873$ & $0.96$ \\
    AutoAug & $6873$ & $0.78$ & $3043$ & $0.66$ & $5033$ & $0.74$ \\
    RandAug & $1335$ & $0.62$ & $4627$ & $0.77$ & 11850 & $0.94$ \\
    Flip  & $187$ & $0.55$ & $6109$ & $0.85$ & $6439$ & $0.85$ \\
    Rotate  & $4223$ & $0.82$ & $9269$ & $0.97$ & $13760$ & $0.92$ \\
    Permute  & $14262$ & $0.97$ & $5561$ & $0.84$ & $12209$ & $0.96$ \\
    MixUp  & $23242$ & $0.72$ & $21107$ & $0.76$ & $19799$ & $0.79$ \\
    FGSM & $28360$ & $0.98$ & $43893$ & $1.00$ & $50337$ & $1.00$ \\
    PGD  & $54513$ & $1.00$ & $93465$ & $1.00$ & $109530$ & $1.00$ \\
    \bottomrule
    \end{tabular}%
\end{sc}
\end{small}
\end{center}
\vskip -0.2in
\end{table*}

\end{document}